\documentclass[sigconf]{acmart}

\AtBeginDocument{%
  }

\copyrightyear{2026}
\acmYear{2026}
\setcopyright{cc}
\setcctype{by}
\acmConference[MM '26]{Proceedings of the 34th ACM International Conference on Multimedia}{November 10--14, 2026}{Rio de Janeiro, Brazil}
\acmBooktitle{Proceedings of the 34th ACM International Conference on Multimedia (MM '26), November 10--14, 2026, Rio de Janeiro, Brazil}
\acmDOI{10.1145/3767308.3836271}
\acmISBN{979-8-4007-2213-4/2026/11}
\usepackage{placeins}
\usepackage{algorithm}
\usepackage{algpseudocode}
\usepackage{hyphenat}
\usepackage{enumitem}
\usepackage{graphicx}
\usepackage{adjustbox}
\usepackage{tabularx}
\usepackage{multirow}
\usepackage{colortbl}
\usepackage{xcolor}
\usepackage{booktabs}
\usepackage{subfigure}
\usepackage{todonotes}
\usepackage{balance}
\definecolor{oursrow}{HTML}{CBE4FB}

\makeatletter
\def\blfootnote{\gdef\@thefnmark{}\@footnotetext}

\begin{document}

\newcommand{\model}{DualG-MRAG}
\title{\model: Decoupling Macro-Reasoning and Micro-Matching for Multimodal Retrieval-Augmented Generation}

\author{Jiacheng Tao}
\affiliation{%
  \department{SKLCCSE, School of Computer Science and Engineering}
  \institution{Beihang University}
  \city{Beijing}
  \country{China}
}
\email{jiachengtao@buaa.edu.cn}

\author{Qingyun Sun}
\authornote{Corresponding author}
\affiliation{%
  \department{SKLCCSE, School of Computer Science and Engineering}
  \institution{Beihang University}
  \city{Beijing}
  \country{China}
}
\email{sunqy@buaa.edu.cn}

\author{Haonan Yuan}
\affiliation{%
  \department{SKLCCSE, School of Computer Science and Engineering}
  \institution{Beihang University}
  \city{Beijing}
  \country{China}
}
\email{yuanhn@buaa.edu.cn}

\author{Ziwei Zhang}
\affiliation{%
  \department{SKLCCSE, School of Computer Science and Engineering}
  \institution{Beihang University}
  \city{Beijing}
  \country{China}
}
\email{zwzhang@buaa.edu.cn}

\author{Jianxin Li}
\affiliation{%
  \department{SKLCCSE, School of Computer Science and Engineering}
  \institution{Beihang University}
  \city{Beijing}
  \country{China}
}
\email{lijx@buaa.edu.cn}

\renewcommand{\shortauthors}{Jiacheng Tao, Qingyun Sun, Haonan Yuan, Ziwei Zhang, and Jianxin Li}

\begin{abstract}
While Multimodal Retrieval-Augmented Generation (MM-RAG) has shown promising results, it still struggles with complex multi-hop reasoning tasks. 
Existing methods primarily focus on independent instance-level matching, which often fails to capture explicit relationships across modalities and documents. 
Although Graph-enhanced methods introduce structural modeling, they face a fundamental challenge in multimodal scenarios: incorporating fine-grained visual features leads to rapid graph expansion and retrieval noise, whereas coarse-grained representations cause the discarding of critical local evidence. 
To address this dilemma, we propose \textbf{DualG-MRAG}, a \textbf{Dual}-tier framework that introduces a decoupled architecture comprising Macro-reasoning and Micro-matching \textbf{G}raphs for \textbf{M}ultimodal \textbf{RAG}.
Specifically, to suppress retrieval noise by isolating global structural reasoning from fine-grained evidence matching, 
we construct a Macro Graph for global topological routing and a Micro Graph for precise local verification. 
Subsequently, to enable dynamic relevance propagation across heterogeneous evidence sources, we formulate retrieval as a query-driven message passing process via a GNN Retriever.
Furthermore, to provide the generative model with coherent structural guidance, we introduce a dynamic programming decoding mechanism that extracts explicit reasoning paths directly from the GNN's forward pass, replacing the standard input of isolated document chunks. Extensive experiments demonstrate that DualG-MRAG outperforms baselines in both evidence recall and complex QA accuracy.
\end{abstract}

\ccsdesc[500]{Information systems~Question answering}

\keywords{Multimodal Large Language Model, Retrieval-Augmented Generation, Graph Reasoning, Graph Neural Network}

\maketitle
\section{Introduction}
\label{sec:intro}

\begin{figure}[t]
    \centering
    \includegraphics[width=\columnwidth]{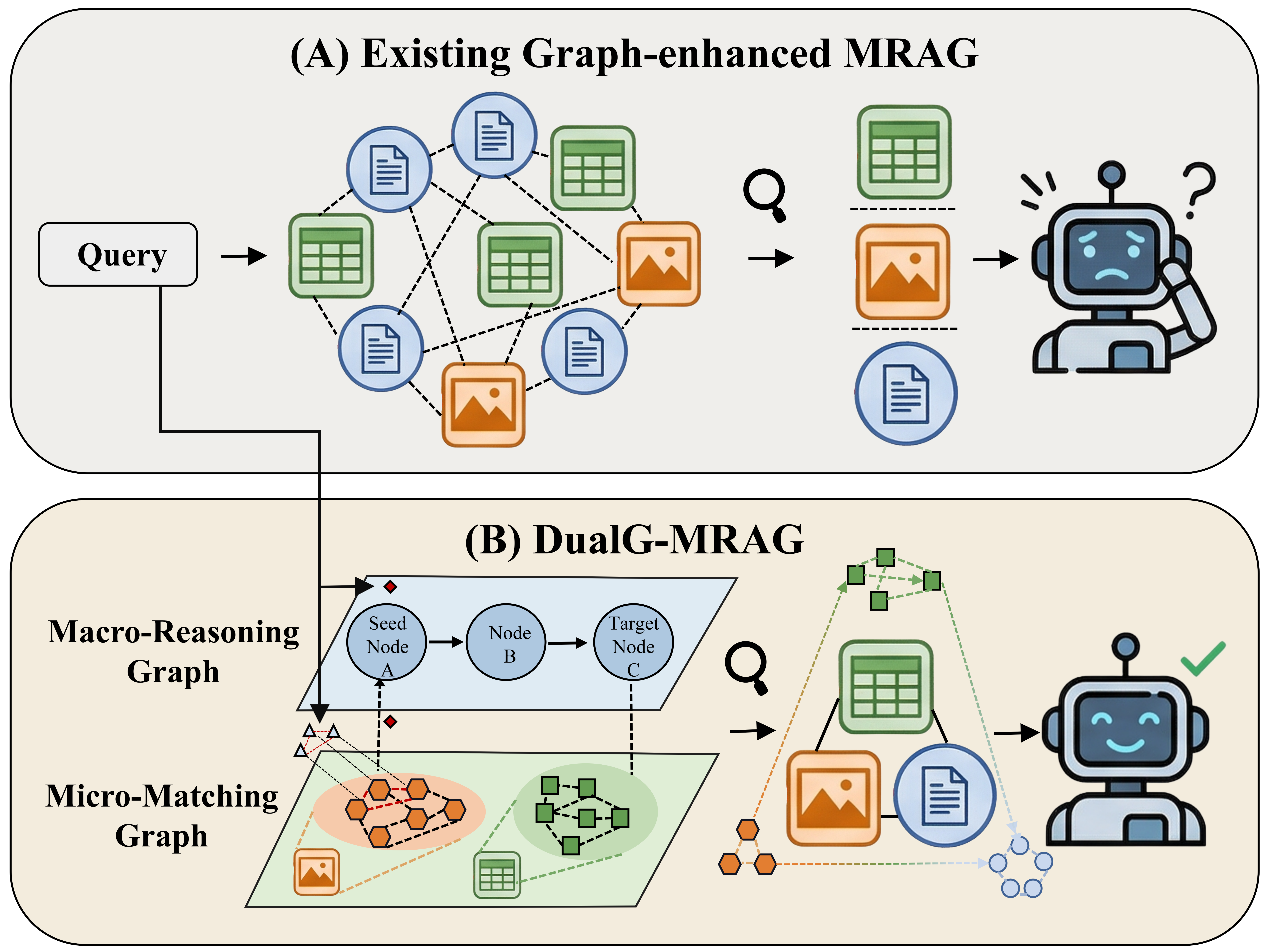}
    \caption{Comparison of existing graph-enhanced MRAG and the proposed DualG-MRAG framework.}
    \label{fig:comparison}
\end{figure}

Multimodal Large Language Models (MLLMs) have achieved remarkable success across various tasks; however, they still struggle with long-context reasoning and are prone to hallucinations when handling knowledge-intensive queries. To enhance their capabilities, MM-RAG extends the traditional RAG framework to heterogeneous data sources, aiming to retrieve and utilize multimodal knowledge in a unified manner~\cite{mei2025survey, abootorabi2025ask}. Despite this progress, existing methods~\cite{radford2021learning, chen2022murag, yu2024visrag, jiang2024vlm2vec, chen2024mllm} mainly rely on coarse-grained alignment paradigms, ranging from dual-encoders to LLM-based embedding frameworks. While effective for general retrieval, these methods often compress complex multimodal evidence into static embeddings, making it difficult to explicitly capture fine-grained dependencies across modalities and documents. Consequently, these methods exhibit clear limitations when handling complex Question Answering (QA) tasks that require multi-hop reasoning.

To address this limitation, recent studies have explored graph-enhanced methods, which improve cross-source information fusion and reasoning capabilities through structural modeling~\cite{wan2025mmgraphrag,park2025m}. Nevertheless, extending such methods to multimodal settings remains challenging. First, in Multimodal Knowledge Graphs, naively incorporating fine-grained features easily leads to a rapid graph expansion and considerable retrieval noise, making it difficult to strike a balance between macro-level reasoning and micro-level matching~\cite{liu2025aligning}. Second, while existing graph-based retrievers are effective at complex relation modeling, their graph structures and information propagation rule are often tailored for textual data only ~\cite{he2024g,luo2025gfm,gutierrez2025rag} or rely on static topologies~\cite{wan2025mmgraphrag,ling2025mmkb}.

To build a graph-enhanced retrieval system that handles multimodal heterogeneous data with both deep reasoning capability and minimized retrieval noise, it is essential to overcome three interconnected core challenges.

    \textbf{Challenge I: Balancing fine-grained representation with the risk of retrieval noise.} 
    Constructing a unified Multimodal Knowledge Graph (MMKG) inherently presents a structural trade-off. 
    On one hand, directly integrating fine-grained visual details into the global topology triggers rapid graph expansion and introduces visual retrieval noise during search. 
    On the other hand, relying solely on coarse-grained, abstract entities discards critical local evidence necessary for precise feature verification. 

    \textbf{Challenge II: The mismatch between static graph structures and the dynamic nature of queries.} 
    Existing graph-based retrievers predominantly rely on predefined, static topologies where structural connectivity and information propagation weights remain largely query-agnostic. In complex multimodal scenarios, however, the relevance of a specific relational pathway is highly dependent on the user's intent. A static retrieval mechanism often leads to unconstrained structural propagation, which blindly expands the search space and introduces irrelevant contexts. 

    \textbf{Challenge III: The lack of explicit structural fusion across heterogeneous evidence.}
    Conventional MM-RAG paradigms typically treat retrieved multimodal documents as a flattened list (e.g., simple concatenation of top-$K$ chunks). These methods require the downstream MLLM to reconstruct complex cross-document relationships from fragmented heterogeneous inputs. This structural disconnect increases the cognitive load on the generative model, constraining its ability to perform multi-hop reasoning. Thus, a critical challenge is how to transform isolated retrieved instances into explicitly connected structural reasoning paths.

To tackle the aforementioned challenges, we propose \textbf{DualG-MRAG}, a framework that reformulates multimodal RAG as a structured reasoning process. 
Our method systematically addresses the established hurdles through a cohesive pipeline: we first introduce a \textit{decoupled dual-tier graph} to resolve the structural dilemma (\textbf{Challenge I}) by isolating global routing from local visual verification. Operating on this graph, a \textit{query-driven GNN retriever} overcomes static limitations (\textbf{Challenge II}) to enable dynamic, query-conditioned evidence gathering. Ultimately, to achieve explicit structural fusion (\textbf{Challenge III}), we design a \textit{path decoding mechanism} that extracts coherent reasoning chains to structurally guide the downstream MLLM. By transforming fragmented retrieved instances into transparent, verifiable pathways, our architecture reduces the implicit reasoning burden on the generative model. Our main contributions are summarized as follows:

\begin{itemize}[leftmargin=1.3em]
    \item We introduce a dual-tier multimodal RAG framework that decouples macro-topological routing from micro-feature verification, effectively mitigating visual retrieval noise while preserving essential connectivity for multi-hop reasoning.
    \item For the MM-RAG scenario, we introduce a query-driven Graph Neural Network mechanism for the first time. By enabling query-guided message passing, it dynamically aligns heterogeneous cross-modal evidence and explicitly extracts logical reasoning chains, reducing the cognitive load on downstream MLLMs.
    \item Extensive experiments on multimodal multi-hop reasoning benchmarks demonstrate our proposed architecture improves both document recall and final QA accuracy against baselines.
\end{itemize}

\begin{figure*}[t]
    \centering
    \includegraphics[width=\textwidth]{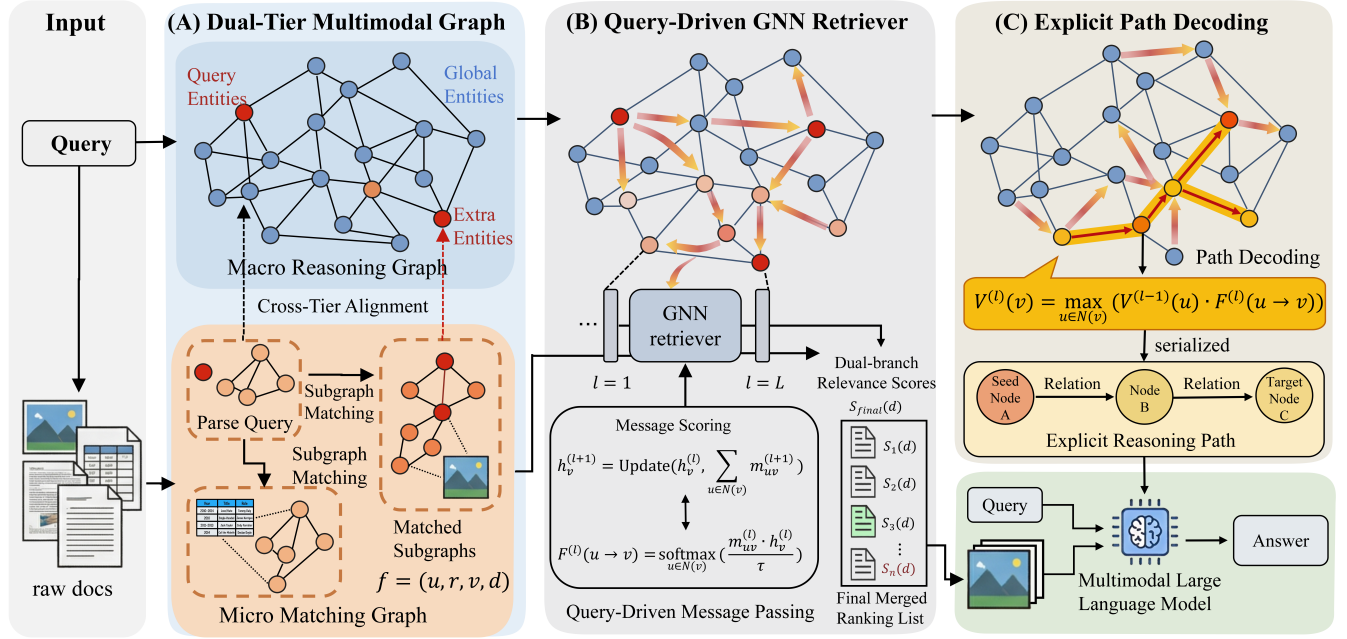}
    \caption{\textbf{The overall architecture of DualG-MRAG.} The framework operates in three query-driven phases. \textbf{(A) Dual-Tier Multimodal Graph:} The query performs graph-level and node-level matching on the Micro Graph. \textbf{(B) Query-Driven GNN Retriever:} The GNN initializes dynamic states from the combined query and supplementary entities, and performs directed message passing across the Macro Graph, yielding the final document relevance scores. \textbf{(C) Explicit Path Decoding:} A layer-wise dynamic programming algorithm extracts optimal reasoning paths directly from the GNN's forward pass, serializing multimodal topological connections to structurally guide the MLLM generation.}
    \label{fig:architecture}
\end{figure*}

\section{Related Work}
\label{sec:related}

\subsection{Multimodal Retrieval-Augmented Generation}
\label{sec:related_mmrag}

Mainstream Multimodal RAG frameworks have evolved by integrating cross-modal retrieval into the generative pipeline. Techniques such as cross-modal contrastive learning and joint representation spaces have been widely adopted to bridge the semantic gap across modalities~\cite{radford2021learning,chen2024mllm,faysse2024colpali}. These methods have shown strong effectiveness in mapping multimodal information into a shared continuous space to facilitate dense retrieval. However, despite achieving high-quality representation alignment, these vector-matching paradigms still struggle to capture explicit relational dependencies. The reliance on proximity-based matching in latent spaces means that while semantically related items are retrieved, the structural connections between them (e.g., how a visual entity precisely grounds a factual claim in a table) remain implicit. This structural limit limits the efficacy of tasks requiring multi-hop reasoning and cross-source evidence integration.

\subsection{Knowledge Graph-Enhanced Multimodal Retrieval}
\label{sec:related_kg}

To provide the structural grounding that latent embeddings lack, researchers have introduced Knowledge Graphs (KGs) to represent explicit entities and relations~\cite{edge2024local,guo2024lightrag,peng2025graph}. When extended to multimodal settings, recent frameworks like RAG-Anything~\cite{guo2025rag} and MMGraphRAG~\cite{wan2025mmgraphrag} attempt to bridge vision and language by constructing multimodal graphs. 
However, these systems encounter challenges in balancing representation granularity with retrieval efficiency. First, mapping fine-grained image patches can lead to a substantial increase in graph scale and computational complexity, whereas relying on coarse-grained entities results in critical multimodal information loss~\cite{yang2025omgm}. Second, their retrieval mechanisms predominantly rely on predefined graph topologies where the information propagation rules do not adapt to the specific reasoning requirements of different user queries. This suggests a potential for architectures that can decouple macro-reasoning from micro-evidence localization.

\subsection{Graph Neural Networks for Retrieval}
\label{sec:related_gnn}

GNNs have proven effective in extracting salient evidence from complex topological dependencies. Recent studies have transitioned from using GNNs merely for node classification to employing them as dense neural retrievers. For instance, GNN-RAG~\cite{mavromatis2024gnn} employs GNNs to reason over candidate nodes, while GFM-RAG~\cite{luo2025gfm} explores graph foundation models to capture global relationship patterns. Nevertheless, most existing GNN-based RAG~\cite{mavromatis2024gnn,luo2025gfm,gutierrez2025rag,luo2025g,yuan2026retrieving} methods primarily focus on textual data, with limited support for the heterogeneous structures of images and tables. 
Furthermore, despite advancements in GNN-based RAG, the outputs of these models
are typically treated as isolated node scores, leaving the intricate multi-hop reasoning paths implicit. They lack a systematic mechanism to extract and integrate heterogeneous evidence into explicit, readable reasoning paths. 
Consequently, establishing a systematic mechanism to integrate heterogeneous evidence into explicit reasoning paths remains a challenge.

\section{Preliminaries}
\label{sec:preliminaries}

A Knowledge Graph (KG) is formally defined as a directed relational graph $\mathcal{G} = (\mathcal{V}, \mathcal{R})$, with entities $\mathcal{V}$ and relations $\mathcal{R}$. Its fundamental unit is a factual triple $\mathcal{T} = \{(h, r, t) \mid h, t \in \mathcal{V}, r \in \mathcal{R}\}$. 

Given a user query $q$ and a massive multimodal corpus $\mathcal{D} = \{d_1, \dots, d_N\}$ comprising heterogeneous data (text, images, tables), a standard MM-RAG system retrieves a relevant evidence subset $\mathcal{D}_{\text{ret}} \subset \mathcal{D}$. A Multimodal Large Language Model (MLLM), parameterized by $\theta$, then autoregressively generates the answer $A$ by maximizing the conditional probability:
$$
    P(A \mid q, \mathcal{D}_{\text{ret}}) = \prod_{t=1}^{|A|} P_\theta(a_t \mid a_{<t}, q, \mathcal{D}_{\text{ret}}).
$$

\section{Framework: DualG-MRAG}
\label{sec:method}

We propose a framework that decouples multimodal knowledge into a Macro Graph for cross-document topology and a Micro Graph for intra-document verification. 
As illustrated in Figure~\ref{fig:architecture}, after offline construction (Section~\ref{sec:dual_tier}), inference proceeds sequentially. First, query-driven subgraph matching on the Micro Graph identifies supplementary entities (Section~\ref{sec:matching}). These, alongside query entities, initialize a query-driven GNN retriever that propagates relevance across the Macro Graph to rank candidate documents (Section~\ref{sec:gnn_retriever}). Finally, we decode explicit reasoning paths directly from the GNN's forward pass to structurally guide the downstream MLLM generation (Section~\ref{sec:path_injection}).

\subsection{Dual-Tier Multimodal Graph Construction}
\label{sec:dual_tier}

Rather than forcing heterogeneous multimodal data into a single flattened graph, which often struggles to balance information retention with retrieval efficiency.
We construct a dual-tier Multimodal Knowledge Graph (MMKG).

\subsubsection{\textbf{Macro Reasoning Graph ($\mathcal{T}^M$).}}
The Macro Graph provides a high-level structural backbone. We define $\mathcal{T}^M = (\mathcal{E}^M, \mathcal{R}^M)$, where $\mathcal{E}^M$ denotes the set of global entities, and $\mathcal{R}^M$ represents the set of relations connecting them. To incorporate visual information into the shared reasoning space, we generate concise factual captions for images using a frozen Vision-Language Model (VLM)~\cite{bai2025qwen3} and merge them into the text content. Following Open Information Extraction (OpenIE)~\cite{angeli2015leveraging,pai2024survey,zhou2022survey}, we obtain the base triples $\mathcal{T}^M_{\text{raw}}$, where each triple is structured as $(e_{\text{head}}, r, e_{\text{tail}})$. The connectivity is further enhanced by introducing equivalence edges $\mathcal{T}^{\text{eq}}$ based on semantic entity resolution. Specifically, we compute the cosine similarity between the embeddings of extracted entities using a pre-trained encoder; if the similarity exceeds a predefined threshold $\tau$, an equivalence edge is added to connect them:
\begin{equation}
    \mathcal{T}^M = \mathcal{T}^M_{\text{raw}} \cup \mathcal{T}^{\text{eq}}.
    \label{eq:macro_graph}
\end{equation}

\subsubsection{\textbf{Micro Matching Graph ($\mathcal{T}^m$).}}
In contrast, the Micro Graph is designed to capture fine-grained dependencies within specific images and tables. To explicitly represent these local structures, a micro-fact $f \in \mathcal{T}^m$ is defined as a 4-tuple: $f = (u, r, v, d)$, where $u$ and $v$ are the head and tail nodes, $r$ is the relation, and $d$ is the source document pointer (e.g., image file paths).

We represent the head and tail nodes $u$ and $v$ as textual anchors, and this abstraction does not result in the loss of fine-grained visual features. The rich visual details, such as spatial relationships, object attributes, and local interactions, are explicitly captured into the specific relations $r$ of the micro-triples. Furthermore, the pointer $d$ explicitly links these micro-facts back to the original raw images.
The raw images or tables referenced by $d$ are directly fed into the downstream MLLM alongside the extracted reasoning paths.

\subsection{Structural Matching and Evidence Fusion}
\label{sec:matching}

To enable the system to perform multi-hop reasoning grounded in fine-grained facts, we establish a dynamic activation mechanism. 

\subsubsection{\textbf{Cross-Tier Alignment.}}
We define an explicit mapping function $C(\cdot): \mathcal{E}^m \to \mathcal{E}^M$, where $\mathcal{E}^m$ denotes the set of nodes within the Micro Graph, to align micro-facts with global entities. For any node $e$ in the Micro Graph, we employ a two-stage linking strategy: exact string matching followed by soft semantic linking via a pre-trained ColBERT encoder~\cite{khattab2020colbert}. This ensures that localized multimodal concepts are anchored to the global topology.

\subsubsection{\textbf{Dual-Branch Evidence Activation.}}
To robustly extract evidence from the Micro Graph, we employ a dual-branch activation strategy: a rigorous graph-level subgraph matching pipeline and an auxiliary node-level retrieval branch. 

\textbf{Graph-Level Structural Matching.}
During retrieval, an input query $q$ is first processed by a constrained LLM-based parser to generate a structured pattern graph $P(q)$:
\begin{equation}
    P(q) = (\mathcal{T}_{\text{cond}}(q), \mathcal{T}_{\text{target}}(q), k_v(q)).
\end{equation}
Here, $\mathcal{T}_{\text{cond}}$ represents explicit evidence constraints (formalized as complete relational triples with optional wildcards), and $\mathcal{T}_{\text{target}}$ specifies the exact target entity or attribute required to answer the query. Additionally, $k_v(q)$ acts as a dynamic visual budget. This integer explicitly quantifies the query's modality preference and regulates the maximum number of raw images ultimately fed to the downstream MLLM, preventing visual context overload for text-centric questions. To ground the reasoning, we perform subgraph matching~\cite{cai2025simgrag} on the Micro Graph $\mathcal{T}^m$ driven by the constraint graph $\mathcal{T}_{\text{cond}}$. For a candidate mapping $\pi$, the matching cost is computed as the sum of joint semantic distance of nodes and relations:
\begin{equation}
    \text{dist}(\pi) = \sum_{(u,r,v) \in \mathcal{T}_{\text{cond}}} d_n(u, \pi(u)) + d_r(r, \pi(r)) + d_n(v, \pi(v)),
    \label{eq:matching_cost}
\end{equation}
where $d_n(\cdot, \cdot)$ and $d_r(\cdot, \cdot)$ denote the distance functions between the query elements and the mapped micro-graph candidates.

Since exact subgraph isomorphism is inherently NP-hard, exhaustive search over a massive micro-graph is computationally prohibitive~\cite{ullmann1976algorithm}. To maintain acceptable retrieval latency, we utilize an approximate heuristic matching pipeline. Specifically, we first restrict the search space by retrieving only Top-$K$ semantic candidates for the query nodes and relations via dense vector indexing. Within this reduced subspace, we execute a heuristic Branch-and-Bound search. By employing topology-aware traversal and hard semantic gating to prune unpromising branches early, we efficiently approximate the optimal mapping. 

We retain the Top-$K$ subgraphs with the minimum $\text{dist}(\pi)$. Utilizing the alignment index $C$, the micro-facts explicitly hit by $\mathcal{T}_{\text{cond}}$ are projected into the Macro Graph, forming an extended activation set $\mathcal{S}_{\text{ext}}(q)$. These nodes are merged with standard Named Entity Recognition (NER)~\cite{keraghel2024recent} results to produce the final query mask $\mathbf{m}_q$. Instead of a static vector search, this mask $\mathbf{m}_q$ serves as the initialization input for our subsequent GNN retriever. Concurrently, this structural match yields a graph-level document score $s_{\text{graph}}(d)$ for the source documents.

\textbf{Node-Level Document Boosting.}
Relying solely on strict subgraph matching is vulnerable to parser failures or incomplete graph extractions. To address this, we introduce an auxiliary node-level matching branch. We utilize explicit entities mentioned in $q$ as anchors to perform a nearest-neighbor search within the micro-node space. However, we impose a strict boundary: the matched nodes from this branch do not expand the GNN seed mask $\mathbf{m}_q$. Instead, they solely contribute to a node-level document score $s_{\text{node}}(d)$.

\textbf{Hybrid Evidence Fusion.}
Finally, the scores from both pathways are merged to form the overall micro-matching score for document retrieval:
\begin{equation}
    s_{\text{micro}}(d) = \max\Bigl(s_{\text{graph}}(d), s_{\text{node}}(d)\Bigr).
    \label{eq:hybrid_fusion}
\end{equation}

\subsection{Query-aware GNN Retriever}
\label{sec:gnn_retriever}

Traditional multimodal retrieval systems rely on static dense vector similarity, which struggles to capture the intricate, multi-hop dependencies between entities. To overcome this limitation, we formulate retrieval as a dynamic, query-driven message passing mechanism~\cite{galkin2023towards,zhu2021neural} over the Macro Graph $\mathcal{T}^M$, employing the NBFNet architecture~\cite{zhu2021neural,luo2025gfm} as our architectural backbone.

\subsubsection{\textbf{Dynamic State Initialization.}}
Unlike static Graph Neural Networks (GNNs)~\cite{kipf2016semi,gilmer2017neural} that utilize fixed structural node embeddings, making the graph's message passing process agnostic to the user's intent, our model initializes the hidden state dynamically. Let $\mathbf{m}_q$ be the binary mask of activated macro-nodes identified during the cross-tier alignment (Section~\ref{sec:matching}). For each node $v \in \mathcal{E}^M$, the initial hidden state $h_v^{(0)}$ is defined by injecting the semantic embedding of the query $q$:
\begin{equation}
    h_v^{(0)} = \begin{cases} \text{Enc}(q), & \text{if } v \in \mathbf{m}_{q}, \\ \mathbf{0}, & \text{otherwise,} \end{cases}
    \label{eq:init_state}
\end{equation}
where $\text{Enc}(\cdot)$ is a pre-trained all-mpnet-v2 text encoder.

Intuitively, this initialization mechanism ensures that the neural information flow originates exclusively from query-relevant anchors, effectively pruning the vast, noisy search space of the MMKG at the initialization stage.

\subsubsection{\textbf{Relational Message Passing.}}
To model the semantic evolution across multiple reasoning hops, we perform $L$ layers of message passing. At the $(l+1)$-th layer, for every edge $(u, r, v) \in \mathcal{T}^M$, the message $m_{uv}^{(l+1)}$ integrates the source node state, the relation projection, and the target node state:
\begin{equation}
    m_{uv}^{(l+1)} = \text{Msg}\left(h_u^{(l)},\; g^{(l+1)}(h_r),\; h_v^{(l)}\right).
    \label{eq:message}
\end{equation}
Specifically, the message function $\text{Msg}(\cdot)$ is implemented using a non-parametric DistMult operation~\cite{yang2014embedding}. Here, $g^{(l+1)}(\cdot)$ is a layer-specific relation transformation that allows the model to learn distinct traversal logic at varying reasoning depths. The node state is subsequently updated by aggregating messages from its topological neighborhood $\mathcal{N}(v)$:
\begin{equation}
    h_v^{(l+1)} = \text{Update}\left(h_v^{(l)},\; \sum_{u \in \mathcal{N}(v)} m_{uv}^{(l+1)}\right).
    \label{eq:update}
\end{equation}
We instantiate the $\text{Update}$ function by first aggregating the incoming messages via sum pooling, followed by a single linear transformation to update the node's representation.

\subsubsection{\textbf{Relevance Scoring and Document Fusion.}}
After $L$ layers of propagation, the final hidden state $h_v^{(L)}$ encapsulates the multi-hop topological relevance of node $v$ relative to $q$. We predict a relevance score $P_q(v)$ for each node via a Multi-Layer Perceptron~\cite{luo2025gfm}:
\begin{equation}
    P_q(v) = \text{MLP}(h_v^{(L)}).
    \label{eq:score}
\end{equation}

Since the macro-entities $\mathcal{E}^M$ serve as shared structural anchors connecting text and visual concepts, we project these node-level scores back to the document space via sparse matrix multiplication, yielding raw textual/tabular scores $t_d$ and visual scores $i_d$. To ensure fair cross-modal fusion, we independently apply Min-Max normalization to obtain $\hat{t}_d$ and $\hat{i}_d$. 

To mitigate unverified visual noise, the visual scores are explicitly modulated by the structural micro-matching score $s_{\text{micro}}(d)$. For candidate documents that hit the micro-constraints, their visual scores are updated via a weighted addition of the normalized visual score and the micro-score (i.e., $\alpha \cdot \hat{i}_d + \beta \cdot s_{\text{micro}}(d)$), followed by re-normalization. For documents lacking micro-evidence, their visual scores are simply multiplied by a decay scale. Let the resulting modulated visual score be $\hat{i}^*_d$. Finally, the final ranking score for a multimodal document $d$ is determined by a max operation:
\begin{equation}
    s_{\text{final}}(d) = \max(\hat{t}_d, \hat{i}^*_d).
    \label{eq:final_merge}
\end{equation}

Through this formulation, documents are highly ranked if they possess strong textual reasoning or structurally verified visual evidence.

\begin{table*}[t]
\centering
\setlength{\tabcolsep}{2.5mm}
\caption{Main experimental results on MMQA and WebQA datasets. Performance is evaluated using Exact Match (EM) and F1 for MMQA, and ROUGE-L (R-L) and BERTScore (BERTSc.) for WebQA. All results are reported in percentages (\%). The best results are highlighted in \textbf{bold}, and the second-best results are \underline{underlined}.}
\label{tab:main_results}
\begin{tabular}{ll|cccc|cccc}
\toprule
\multirow{2}[6]{*}{\textbf{Category}} & \multirow{2}[6]{*}{\textbf{Method}} & \multicolumn{4}{c|}{\textbf{Qwen3-VL-4B}} & \multicolumn{4}{c}{\textbf{Qwen3-VL-8B}} \\
\cmidrule(lr){3-6} \cmidrule(lr){7-10}
 &  & \multicolumn{2}{c}{\textbf{MMQA}} & \multicolumn{2}{c|}{\textbf{WebQA}} & \multicolumn{2}{c}{\textbf{MMQA}} & \multicolumn{2}{c}{\textbf{WebQA}} \\
 &  & EM & F1 & R-L & BERTSc. & EM & F1 & R-L & BERTSc. \\
\midrule
Base LLM & None & 18.80 & 21.18 & 45.16 & 67.64 & 22.30 & 25.11 & 44.13 & 67.07 \\
\midrule
\multirow{4}{*}{Multimodal RAG} 
 & VisRAG~\scalebox{0.68}{\textcolor{gray}{\textit{ICLR'25}}}~\cite{yu2024visrag} & 27.80 & 30.33 & 47.25 & 69.14 & 31.20 & 34.08 & 45.75 & 67.72 \\
 & VLM2Vec-V2.0~\scalebox{0.68}{\textcolor{gray}{\textit{TMLR'26}}}~\cite{meng2025vlm2vec} & 31.10 & 34.88 & 48.31 & 69.73 & 33.00 & 36.78 & 47.39 & 69.10 \\
 & CoRe-MMRAG~\scalebox{0.68}{\textcolor{gray}{\textit{ACL'25}}}~\cite{tian2025core} & 30.60 & 34.02 & 47.16 & 65.72 & 35.00 & 39.15 & 45.85 & 65.63 \\
 & ViDoRAG~\scalebox{0.68}{\textcolor{gray}{\textit{EMNLP'25}}}~\cite{wang2025vidorag} & \underline{37.20} & \underline{41.77} & 48.54 & 66.03 & \underline{40.00} & \underline{43.76} & 47.13 & 66.06 \\
\midrule
\multirow{3}{*}{Graph-enhanced RAG} 
 & HM-RAG~\scalebox{0.68}{\textcolor{gray}{\textit{MM'25}}}~\cite{liu2025hm} & 34.30 & 39.05 & 45.44 & 67.96 & 35.90 & 41.36 & 44.26 & 67.70 \\
 & MMGraphRAG~\scalebox{0.68}{\textcolor{gray}{\textit{AAAI'26}}}~\cite{wan2025mmgraphrag} & 35.50 & 39.42 & \underline{48.60} & \underline{70.08} & 39.50 & 43.32 & \underline{47.96} & \underline{69.41} \\

 & \cellcolor{gray!15} \textbf{DualG-MRAG (Ours)} & \cellcolor{gray!15} \textbf{44.20} & \cellcolor{gray!15} \textbf{47.57} & \cellcolor{gray!15} \textbf{50.10} & \cellcolor{gray!15} \textbf{70.58} & \cellcolor{gray!15} \textbf{46.00} & \cellcolor{gray!15} \textbf{51.19} & \cellcolor{gray!15} \textbf{48.92} & \cellcolor{gray!15} \textbf{69.74} \\
\bottomrule
\end{tabular}
\end{table*}

\subsection{Explicit Path Injection for Evidence Fusion}
\label{sec:path_injection}

Conventional Multimodal RAG systems typically treat retrieved heterogeneous documents as an isolated, flattened list. This forces the downstream Multimodal Large Language Model (MLLM) to implicitly infer the latent cross-document relationships during generation. We argue that the \textbf{topological connectivity} between evidence, such as how a visual entity in an image structurally leads to a factual cell in a table, is just as informative as the evidence itself.
Building upon the path-based retrieval explored in PathRAG~\cite{chen2026pathrag}, which utilizes heuristic search to identify relevant contexts, we shift the focus toward an endogenous approach. Instead of relying on external search heuristics or explicit probabilistic modeling~\cite{li2026bayesrag}, we recover these explicit reasoning paths directly by tracking the message passing trajectories from the GNN's forward pass.

\subsubsection{\textbf{Local Flow Decomposition.}}
During the message passing phase (Section~\ref{sec:gnn_retriever}), the edge message $m_{uv}^{(l)}$ is designed to capture the structural influence of node $u$ on node $v$. We quantify this contribution by defining a Local Flow Probability $F^{(l)}(u \to v)$. This is computed via the scaled dot-product between the incoming message and the target node's updated state:
\begin{equation}
    F^{(l)}(u \to v) = \text{softmax}_{u \in \mathcal{N}(v)} \left( \frac{m_{uv}^{(l)} \cdot h_v^{(l)}}{\tau} \right),
    \label{eq:local_flow}
\end{equation}
where $\tau$ is a temperature hyperparameter controlling the sparsity of the flow distribution. A lower $\tau$ encourages the network to concentrate on a few dominant reasoning paths rather than diffusing energy uniformly.

\subsubsection{\textbf{Efficient Path Decoding via Dynamic Programming.}}
To provide the MLLM with a coherent evidence sequence, we extract the optimal evidence chains using a layer-wise dynamic programming algorithm over the flow probabilities. Although the original KG naturally contains cycles, the $L$ layers of message passing can be naturally modeled as an $L$-hop computational Directed Acyclic Graph (DAG). Since the transition to state $v$ at hop $l$ strictly depends on the states at hop $l-1$, finding the most probable path reduces to a highly efficient Dynamic Programming (DP) process on this DAG. Let $V^{(l)}(v)$ denote the maximum cumulative path probability reaching node $v$ at hop $l$. We recursively compute:
\begin{equation}
    V^{(l)}(v) = \max_{u \in \mathcal{N}(v)} \left( V^{(l-1)}(u) \cdot F^{(l)}(u \to v) \right).
    \label{eq:viterbi}
\end{equation}

By tracking the optimal predecessor $\Psi^{(l)}(v) = \arg\max_{u \in \mathcal{N}(v)} (\cdot)$, we can backtrack from the highest-scoring target nodes (obtained in Section~\ref{sec:gnn_retriever}) to the initial query entities. 
Since the decoding is performed exclusively on the restricted $L$-hop computational subgraph rather than the massive original KG, the computational overhead is accordingly limited.

\subsubsection{\textbf{Topological Evidence Serialization.}}
The output generated by path decoding forms a set of explicit reasoning path sequences: $\mathcal{P} = \{(e_0, r_1, e_1, \dots, e_L)\}$, where $e_i \in \mathcal{E}^M$ represents the sequence of global macro-entities and $r_i \in \mathcal{R}^M$ denotes the intermediate relations connecting them at each hop. Instead of feeding the MLLM an unorganized collection of retrieved chunks, we serialize these optimal paths into a \textit{Structured Evidence Graph}.

Specifically, we utilize the cross-tier alignment index $C(\cdot)$ to inversely map the abstract macro-entities within the decoded path back to their original multimodal sources. These linked multi-source snippets spanning textual paragraphs, image regions, and tabular data are subsequently explicitly verbalized as step-by-step reasoning paths in the generation prompt.

\section{Experiments}
\label{sec:exp}

In this section, we conduct extensive experiments to address the following research questions:

\begin{itemize}[leftmargin=*]
    \item \textbf{RQ1:} How does \model{} perform on complex multi-hop QA tasks?
    \item \textbf{RQ2:} Does the model exhibit strong cross-domain robustness across different subjects, difficulty levels, and heterogeneous modality contexts?
    \item \textbf{RQ3:} Can the macro-micro decoupled architecture effectively improve retrieval?
    \item \textbf{RQ4:} Can the method achieve a trade-off between retrieval efficiency and QA performance?
    \item \textbf{RQ5:} What are the specific contributions of each core component to the final system performance?
\end{itemize}

\begin{table*}[htbp]
\centering
\setlength{\tabcolsep}{3.9mm}
\caption{Fine-grained performance breakdown (Accuracy \%) on the ScienceQA dataset. We evaluate models across different Subjects, Context Modalities, and Grade levels. The best results among computational models are highlighted in \textbf{bold}, and the second-best are \underline{underlined}. ``Human'' and ``GPT-4 (CoT)'' serve as reference anchors.}
\label{tab:sciqa_results}
\begin{tabular}{l|ccc|ccc|cc|c}
\toprule
\multirow{2}[2]{*}{\textbf{Method}} & \multicolumn{3}{c|}{\textbf{Subject}} & \multicolumn{3}{c|}{\textbf{Context Modality}} & \multicolumn{2}{c|}{\textbf{Grade}} & \multirow{2}[2]{*}{\textbf{Avg.}} \\
\cmidrule(lr){2-4} \cmidrule(lr){5-7} \cmidrule(lr){8-9}
 & NAT & SOC & LAN & TXT & IMG & NO & G1-6 & G7-12 & \\
\midrule
Human Performance & \underline{90.23} & 84.97 & 87.48 & 89.60 & 87.50 & 88.10 & \underline{91.59} & 82.42 & 88.40 \\
GPT-4 (CoT) & 85.48 & 72.44 & \textbf{90.27} & 82.65 & 71.49 & \textbf{92.89} & 86.66 & 79.04 & 83.99 \\
\midrule
Qwen3-VL-8B (Zero-shot) & 87.83 & \underline{95.73} & 85.82 & 85.80 & \underline{90.48} & 88.57 & 91.37 & 84.64 & 88.96 \\
VLM2Vec-V2.0~\scalebox{0.68}{\textcolor{gray}{\textit{TMLR'26}}}~\cite{meng2025vlm2vec} & 87.83 & 90.66 & 86.64 & 85.80 & 88.10 & 89.41 & 89.83 & 85.04 & 88.12 \\
HM-RAG~\scalebox{0.68}{\textcolor{gray}{\textit{MM'25}}}~\cite{liu2025hm} & 89.21 & 92.58 & 85.91 & \textbf{91.00} & 88.84 & 88.29 & 91.19 & \underline{85.23} & \underline{89.06} \\
\rowcolor{gray!15} 
\textbf{DualG-MRAG (Ours)} & \textbf{90.36} & \textbf{96.18} & \underline{88.09} & \underline{90.75} & \textbf{91.52} & \underline{90.38} & \textbf{92.73} & \textbf{87.87} & \textbf{90.99} \\
\bottomrule
\end{tabular}
\end{table*}

\subsection{Experimental Settings}
\label{sec:exp_settings}

\paragraph{\textbf{Datasets.}}
To evaluate the retrieval and reasoning capabilities of our framework in complex multimodal QA scenarios, we conduct experiments on three benchmark datasets specifically designed for multi-hop reasoning~\cite{abootorabi2025ask}: MultiModalQA (MMQA)~\cite{talmor2021multimodalqa}, WebQA~\cite{chang2022webqa}, and ScienceQA~\cite{lu2022learn}.

\begin{itemize}[leftmargin=*]
    \item \textbf{MMQA} focuses on complex cross-modal multi-hop QA, requiring the system to capture multi-hop topological dependencies across heterogeneous tables, images, and text.
    Following the common evaluation paradigm adopted by existing GraphRAG methods for large-scale corpora~\cite{gutierrez2024hipporag,luo2025gfm}, we randomly sample 1,000 queries from its validation set and construct a local knowledge base from the associated documents for evaluation.
    \item \textbf{WebQA} is a large-scale multimodal QA benchmark that tests the system's ability to identify and integrate relevant visual and textual evidence from heterogeneous sources to generate fluent natural language answers. Similar to MMQA, we randomly select 1,000 samples from its validation set for testing.
    \item \textbf{ScienceQA} evaluates comprehensive reasoning performance across diverse scientific subjects.
    Following previous work~\cite{liu2025hm,liu2025aligning}, we use its training set to build the knowledge base and evaluate on 4,241 test samples.
\end{itemize}

\paragraph{\textbf{Baselines.}} 
We compare \model\ with \textbf{eight} baselines from \textbf{three} primary categories.

\begin{itemize}[leftmargin=*]
    \item \textbf{Base MLLMs:} Qwen3-VL-4B and Qwen3-VL-8B.
    \item \textbf{Multimodal RAG:} VisRAG~\cite{yu2024visrag}, VLM2Vec-V2.0~\cite{meng2025vlm2vec}, CoRe-MMRAG~\cite{tian2025core}, and ViDoRAG~\cite{wang2025vidorag}. These represent vector matching paradigms that rely on continuous latent spaces.
    \item \textbf{Graph-enhanced RAG:} HM-RAG~\cite{liu2025hm} and MMGraphRAG~\cite{wan2025mmgraphrag}, which introduce structural modeling to enhance retrieval, serving as direct competitors to our decoupled graph architecture.
\end{itemize}

\paragraph{\textbf{Implementation Details.}} 
During the graph construction phase, we employ Qwen3-VL-8B to uniformly perform OpenIE extraction for the Macro Graph and fine-grained visual feature parsing for the Micro Graph.  In the answer generation phase, all compared baselines and our method use a unified system prompt, with Qwen3-VL-4B and Qwen3-VL-8B serving as the downstream MLLMs. All experiments are conducted on an NVIDIA A100 GPU cluster.

\begin{figure}[t]
    \centering
    \includegraphics[width=0.96\linewidth]{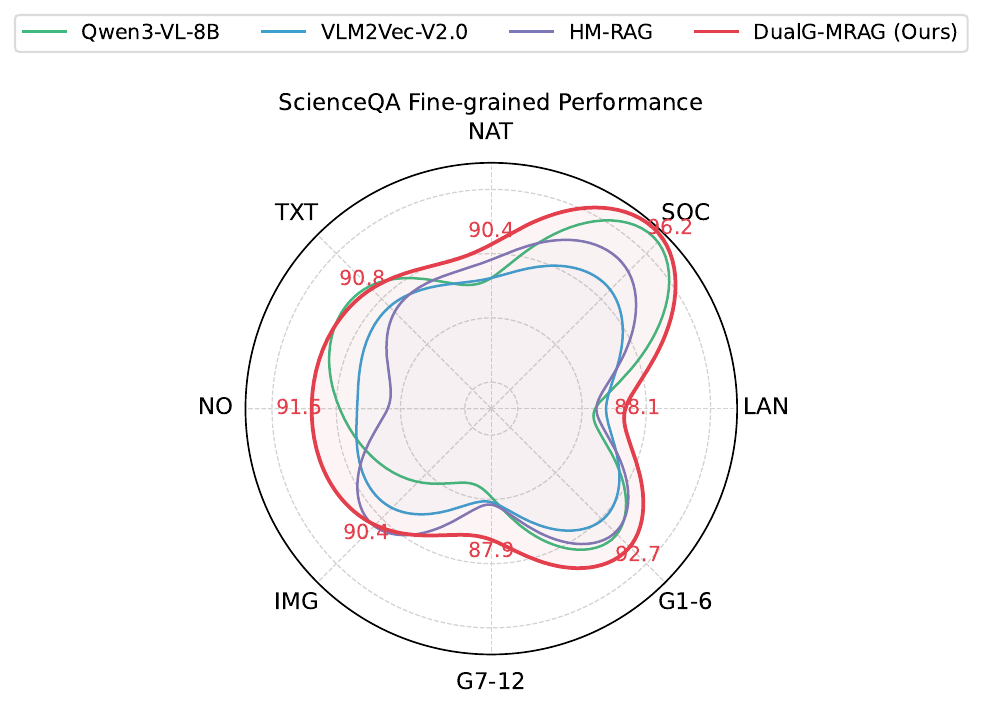}
    \caption{ScienceQA fine-grained performance.}
    \label{fig:radar}
\end{figure}

\subsection{\textit{RQ1:} Performance on QA Tasks}
\label{sec:rq1}

To evaluate the performance of \model{} on complex multi-hop QA tasks, Table~\ref{tab:main_results} presents the end-to-end QA performance of all compared methods on the MMQA and WebQA datasets. The experimental results demonstrate that \model{} surpasses baselines across all evaluation metrics. The key findings are as follows:

\textbf{Effectiveness of Macro-Level Reasoning.}
Traditional multimodal RAG methods are highly dependent on isolated instance-level feature matching, which often limits their capability to integrate cross-document evidence when handling complex multi-hop reasoning tasks. In contrast, \model{} achieves superior performance on the MMQA dataset, achieving an EM score of 44.20\% (with the 4B backbone), which represents an absolute improvement of 7\% over the strongest baseline. This indicates that performing reasoning at the Macro Graph level facilitates the capture of global semantic dependencies better than isolated matching.

\textbf{Advantage of the Decoupled Architecture.}
In multi-hop reasoning scenarios involving massive heterogeneous data such as MMQA and WebQA, existing graph-enhanced methods (e.g., HM-RAG and MMGraphRAG) typically integrate fine-grained visual features directly into a unified graph structure. 
\model{} significantly outperforms these baselines on both datasets. This performance gap suggests that our macro-micro decoupled architecture, which confines micro-feature matching within local nodes rather than global structures, provides a more effective representation for handling complex cross-document reasoning.

\subsection{\textit{RQ2:} Fine-grained Robustness}
\label{sec:rq2}

To investigate the model's fine-grained robustness across different subjects, difficulty levels, and modality contexts, Table~\ref{tab:sciqa_results} (along with the radar chart in Figure~\ref{fig:radar}) presents the fine-grained evaluation results of all models on the ScienceQA dataset. \model{} demonstrates superior cross-modal perception and cross-disciplinary reasoning robustness, achieving an average accuracy of 90.99\%.

\model{} shows strong stability in cross-modal perception.
As shown in the radar chart (Figure~\ref{fig:radar}) and Table~\ref{tab:sciqa_results}, \model{} achieves the best performance of 91.52\% on the IMG subset. This result demonstrates the effectiveness of the Micro Graph.

In the higher-grade problems (G7-12) and Natural Science (NAT) categories that emphasize deep logical deduction, \model{} achieves accuracy of 87.87\% and 90.36\% respectively, widening the performance gap over the best existing graph-enhanced baseline. This indicates that the model can leverage the Macro Graph for reliable multi-hop reasoning when facing complex scientific problems.
\begin{figure*}[t]
    \centering
    \begin{minipage}{\textwidth}
        \centering
        
        \begin{minipage}[b]{0.63\linewidth}
            \centering
            \includegraphics[width=\linewidth]{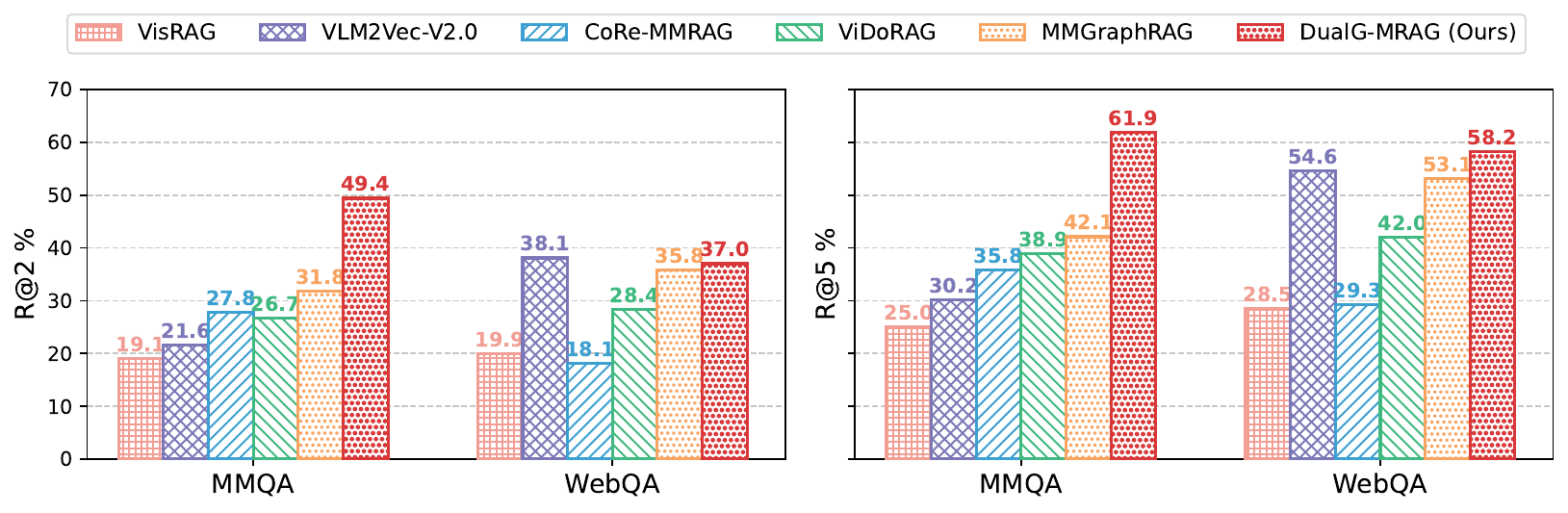} 
            \caption{Top-$K$ Retrieval Recall on MMQA and WebQA datasets.}
            \label{fig:recall}
        \end{minipage}\hfill
        \begin{minipage}[b]{0.365\linewidth}
            \centering
            \includegraphics[width=\linewidth]{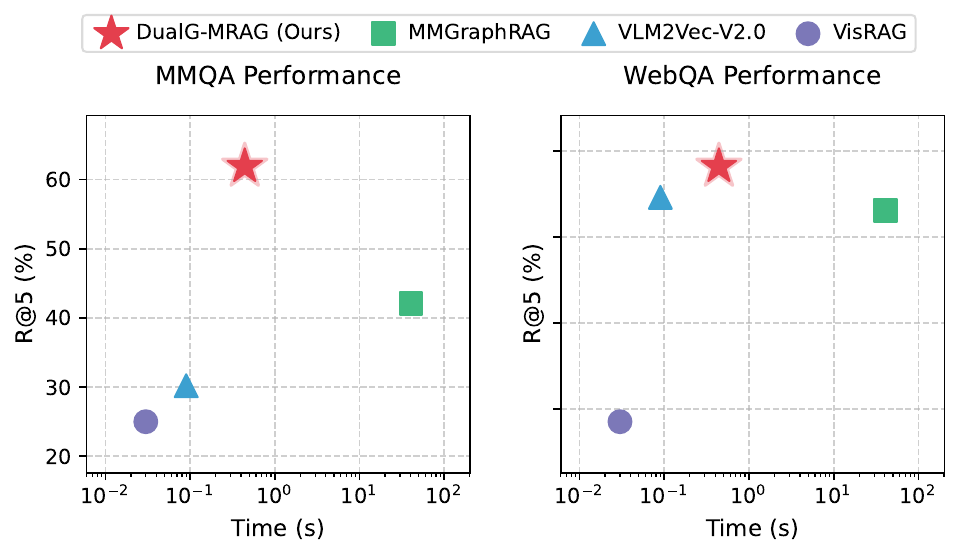} 
            \caption{Efficiency vs.\ Effectiveness.}
            \label{fig:time}
        \end{minipage}
    \end{minipage}
\end{figure*}
\begin{table*}[t]
\centering
\caption{Comprehensive ablation study of DualG-MRAG on MMQA and WebQA. }
\label{tab:comprehensive_ablation}
\setlength{\tabcolsep}{2.5mm}
\begin{tabular}{l|cccc|cccc|cccc}
\toprule
\multirow{3}[5]{*}{\textbf{Architecture Variant}} & \multicolumn{4}{c|}{\textbf{Retrieval Performance}} & \multicolumn{8}{c}{\textbf{Generation Performance}} \\
\cmidrule(lr){2-5} \cmidrule(lr){6-13}
 & \multicolumn{2}{c}{\textbf{MMQA}} & \multicolumn{2}{c|}{\textbf{WebQA}} & \multicolumn{4}{c|}{\textbf{Qwen3-VL-4B}} & \multicolumn{4}{c}{\textbf{Qwen3-VL-8B}} \\
\cmidrule(lr){2-3} \cmidrule(lr){4-5} \cmidrule(lr){6-9} \cmidrule(lr){10-13}
 & R@2 & R@5 & R@2 & R@5 & EM & F1 & R-L & BERTSc. & EM & F1 & R-L & BERTSc. \\
\midrule
\rowcolor{gray!15}
\textbf{DualG-MRAG (Full)} & \textbf{49.4} & \textbf{61.9} & \textbf{37.0} & \textbf{58.2} & \textbf{44.2} & \textbf{47.6} & \textbf{50.1} & \textbf{70.6} & 46.0 & 51.2 & \textbf{48.9} & \textbf{69.7} \\
~\textit{w/o Path Injection} & 49.4 & 61.9 & 37.0 & 58.2 & 41.8 & 46.2 & 48.8 & 69.8 & \textbf{46.5} & \textbf{51.4} & 47.4 & 69.0 \\
~\textit{w/o Micro Graph} & 43.1 & 54.3 & 23.1 & 40.0 & 35.3 & 39.4 & 45.7 & 68.0 & 40.0 & 44.4 & 45.7 & 68.0 \\
~\textit{w/o Macro Graph} & 16.4 & 21.8 & 20.0 & 33.8 & 24.2 & 26.7 & 46.8 & 69.0 & 27.2 & 30.1 & 46.6 & 68.4 \\
\bottomrule
\end{tabular}
\end{table*}

\subsection{\textit{RQ3:} Retrieval Performance}
\label{sec:rq3}

To verify whether the macro-micro decoupled architecture improves retrieval capability, Figure~\ref{fig:recall} presents the Top-$K$ retrieval performance of all methods. The experimental results demonstrate that \model{} can accurately and efficiently capture key supporting evidence from massive heterogeneous document corpora.

On the MMQA dataset, \model{} achieves $R@2$ and $R@5$ of 49.4\% and 61.9\% respectively, marking an improvement over the best graph-enhanced baseline MMGraphRAG (31.8\% and 42.1\%). This result demonstrates the effectiveness of our dual-tier graph architecture in complex retrieval tasks.

On the WebQA dataset, although VLM2Vec-V2.0 slightly leads our method in the $R@2$ metric (38.1\% vs.\ 37.0\%), \model{} quickly overtakes at $R@5$ with a score of 58.2\%. This phenomenon suggests that for long-range complex reasoning tasks, the structured graph often requires a slightly wider retrieval window (e.g., Top-5) to accommodate the complete topological context.

\subsection{\textit{RQ4:} Efficiency and Overhead Analysis}
\label{sec:rq4}

To analyze whether the proposed method achieves a trade-off between retrieval efficiency and QA performance, we evaluate the computational overhead introduced by the dual-tier graph structure. Figure~\ref{fig:time} illustrates the trade-off between retrieval recall and average query latency ($R@5$ vs.\ Time) for each method. In addition, we analyze the internal latency breakdown of \model{} to detail its time distribution. The results indicate that our framework provides a competitive balance between efficiency and effectiveness.

\textbf{Highly Efficient Graph-Enhanced Retrieval Paradigm.}
As observed from the scatter plot in Figure~\ref{fig:time}, existing graph-enhanced methods (e.g., MMGraphRAG) can achieve competitive recall, but suffer from high average latency per query (approximately 40.5 seconds). In contrast, the average query latency of \model{} is $\sim$0.44s. Although this latency remains higher than that of purely lightweight vector matching methods (e.g., VLM2Vec-V2.0 at $\sim$0.09s), \model{} achieves deep reasoning performance gains while maintaining sub-second response speed.

\textbf{Effective Overhead Control via Decoupled Architecture.}
An analysis of the internal latency breakdown reveals that the computational overhead of \model{} is primarily concentrated in Micro-Matching (59.95\%) and Macro-Reasoning (39.40\%).

\subsection{\textit{RQ5:} Ablation Study}
\label{sec:rq5}

We conduct an ablation study to evaluate the specific contributions of DualG-MRAG's core components (Table~\ref{tab:comprehensive_ablation}). We evaluate the impact of Explicit Path Injection (\textit{w/o Path}), Micro-matching Graph (\textit{w/o Micro}), and Macro-reasoning Graph (\textit{w/o Macro}) on both intermediate retrieval recall and final generation quality across two MLLM backbones. Note that Path Injection is a post-retrieval routing strategy, thus its retrieval metrics are identical to the Full model.

\textbf{Indispensability of Macro and Micro Graphs.} 
Removing the Macro Graph (\textit{w/o Macro}) causes $R@5$ on MMQA to plummet from 61.9\% to 21.8\%, proving that global topological connectivity is vital for cross-modal multi-hop routing. Removing the Micro Graph (\textit{w/o Micro}) drops $R@5$ on WebQA by 18.2\%, highlighting its critical role in filtering local visual noise.

\textbf{Interplay between Path Injection and Model Capacity.} 
The impact of Explicit Path Injection varies with the MLLM's scale. For the Qwen3-VL-4B, removing this module degrades performance across all tasks (e.g., a 2.4\% EM drop on MMQA), demonstrating that smaller models rely on explicit structural guidance to process heterogeneous contexts. Conversely, on the more capable Qwen3-VL-8B, removing path injection yields a marginal EM improvement (46.0\% to 46.5\%) on the factoid-oriented MMQA task, suggesting that the path formatting may over-constrain its reasoning flexibility.

\section{Conclusion}
\label{sec:conclusion}

In this paper, we propose \textbf{DualG-MRAG}, a novel dual-tier multimodal RAG framework designed to balance macro reasoning with micro visual verification. By decoupling knowledge representation into a Macro-Reasoning Graph and Micro-Matching Graphs, our method mitigates retrieval noise while preserving essential structural connectivity. 
Furthermore, we formulate the retrieval process as a query-driven message passing mechanism via a GNN, coupled with an explicit path decoding algorithm to provide downstream MLLMs with coherent reasoning chains.
These structured paths reduce the MLLM's implicit reasoning burden.
Extensive experiments 
demonstrate that DualG-MRAG outperforms existing baselines.

\section*{Acknowledgments}
The corresponding author is Qingyun Sun. This work is supported by Beijing Natural Science Foundation under grants No.QY26143, NSFC under grants No.62427808 and No.62225202, and by the Fundamental Research Funds for the Central Universities. We extend our sincere thanks to all reviewers for their valuable efforts.

\clearpage
\bibliographystyle{ACM-Reference-Format}
\balance
\bibliography{ref}

\clearpage
\appendix
\counterwithin{table}{section}
\counterwithin{figure}{section}
\counterwithin{equation}{section}

\section{Algorithm and Complexity Analysis}
\label{sec:appendix_alg}

Constructing a Multimodal Knowledge Graph often leads to an exponential explosion in the number of nodes and edges, especially when fine-grained visual features are incorporated. In this section, we provide a theoretical analysis to demonstrate how \model{} effectively reduces the computational time complexity from a global graph level to a localized subgraph level through its macro-micro decoupled architecture and query-driven dynamic initialization.

\subsection{Time Complexity of Micro-Graph Subgraph Matching}
\label{sec:complexity_micro}

The structural matching pipeline on the Micro Graph $\mathcal{T}^m$ (detailed in Section~\ref{sec:matching}) relies on a two-stage heuristic retrieval process to avoid the NP-hard nature of exact subgraph isomorphism. Let $N_m$ denote the total number of nodes in $\mathcal{T}^m$.

\paragraph{Candidate Anchor Retrieval.} 
Instead of searching the entire graph, we first utilize a pre-trained dense vector index (e.g., FAISS) to retrieve the top-$K$ semantic candidate nodes. This limits the initial search space in sub-linear or logarithmic time, yielding a complexity of $\mathcal{O}(\log N_m)$.

\paragraph{Heuristic Branch-and-Bound Search.} 
Let $k$ be the number of nodes in the query-driven constraint graph $\mathcal{T}_{\text{cond}}$, and $\Delta_m$ be the maximum degree of nodes in $\mathcal{T}^m$. While an exhaustive search would require $\mathcal{O}(N_m^k)$, our method executes a heuristic Branch-and-Bound search strictly within the bounded top-$K$ subspace. By employing hard semantic gating to prune unpromising branches early, the actual branching factor $b$ is significantly smaller than $\Delta_m$ ($b \ll \Delta_m$). Consequently, the worst-case time complexity for this stage is effectively compressed to $\mathcal{O}(K \cdot b^k)$. Given that the query graph size $k$ is typically very small (e.g., $k < 5$) in realistic multi-hop QA scenarios, the computational overhead of this step remains manageable and nearly constant during inference.

\subsection{Time Complexity of Query-Driven Macro-Routing}
\label{sec:complexity_macro}

Traditional Graph Neural Networks (GNNs) execute message passing over the entire graph topology. Let $|\mathcal{E}^M|$ and $|\mathcal{R}^M|$ denote the total number of entities and relations in the Macro Graph $\mathcal{T}^M$, respectively. The conventional per-layer time complexity is $\mathcal{O}(|\mathcal{E}^M| + |\mathcal{R}^M|)$, which is computationally prohibitive for a massive multimodal corpus.

In \model{}, the GNN Retriever is guided by a dynamic state initialization (Section~\ref{sec:gnn_retriever}). The initial activation mask $\mathbf{m}_q$ restricts the neural information flow exclusively to the query-relevant anchors. Let $|\mathcal{E}_{\text{active}}|$ represent the number of active nodes within the localized subgraph after $L$ layers of propagation, where $|\mathcal{E}_{\text{active}}| \ll |\mathcal{E}^M|$. The time complexity of the $L$-layer query-driven message passing is thus bounded by $\mathcal{O}(L \cdot |\mathcal{E}_{\text{active}}| \cdot d^2)$, where $d$ is the hidden state dimension. This dynamic pruning fundamentally circumvents the inefficient global computation over the entire macro-topology.

\subsection{Time Complexity of Explicit Path Decoding}
\label{sec:complexity_path}

To provide structural guidance to the downstream MLLM, we extract optimal reasoning paths directly from the GNN's forward pass (Section~\ref{sec:path_injection}). This process is formulated as a layer-wise Dynamic Programming (DP) algorithm over the flow probabilities. 

Although the original knowledge graph naturally contains complex cycles, the $L$-hop message passing history can be unrolled and modeled as an $L$-hop Directed Acyclic Graph (DAG). Finding the most probable evidence chain is equivalent to computing the Viterbi path on this DAG. Let $\Delta_M$ denote the average degree of nodes in the Macro Graph. Since the dynamic programming is strictly executed over the restricted $L$-hop active computational subgraph rather than the entire MMKG, the time complexity is bounded by $\mathcal{O}(L \cdot |\mathcal{E}_{\text{active}}| \cdot \Delta_M)$. This demonstrates that the explicit path decoding scales linearly with respect to the network depth and the localized subgraph size, ensuring minimal latency overhead during online retrieval.

\section{Extended Information on Baselines and Datasets}
\label{sec:appendix_data}

\subsection{Datasets Details}
\label{sec:appendix_datasets}

To comprehensively evaluate the multi-hop reasoning and multimodal integration capabilities of our proposed \model{}, we conduct experiments on three representative benchmark datasets. The fundamental statistics of the evaluation datasets are summarized in Table~\ref{tab:datasets}.

\begin{table}[h]
\centering
\caption{Statistics of the evaluation datasets. The corpus size denotes the total number of heterogeneous multimodal candidate documents available for retrieval.}
\label{tab:datasets}
\begin{tabular}{llrr}
\toprule
\textbf{Dataset} & \textbf{Context Modality} & \textbf{ Queries} & \textbf{Corpus Size} \\
\midrule
MMQA & Text + Image + Table & 1,000 & 17.6K \\
WebQA & Text + Image & 1,000 & 30.3K \\
ScienceQA & Text + Image & 4,241 & 12.7K \\
\bottomrule
\end{tabular}
\end{table}

\paragraph{\textbf{MultiModalQA (MMQA)}} 
MMQA is a large-scale, challenging question-answering dataset containing 29,918 questions, specifically designed to necessitate joint reasoning across text, tables, and images. Unlike previous datasets where a single modality often suffices, approximately 35.7\% of the questions in MMQA strictly require integrating information from multiple modalities to derive the correct answer. The dataset leverages 16 compositional logic templates (e.g., \textit{INTERSECT}, \textit{COMPARE}) to systematically generate questions that demand complex multi-hop reasoning. 

\paragraph{\textbf{WebQA}} 
WebQA is an open-domain benchmark focusing on multi-hop and multimodal reasoning. It simulates real-world web search scenarios where a system must aggregate knowledge from diverse text snippets and image-caption pairs to generate fluent, natural language answers. A significant characteristic of WebQA is its emphasis on multi-hop reasoning; 44\% of the image-based queries and 99\% of the text-based queries require combining evidence from at least two distinct knowledge sources. 

\paragraph{\textbf{ScienceQA}} 
ScienceQA is an extensive multimodal science question-answering dataset comprising 21,208 multiple-choice questions that span across natural sciences, social sciences, and language sciences. Its high diversity covers 26 topics and 379 specific skill sets across elementary to high school levels (K-12).

\subsection{Baselines Details}
\label{sec:appendix_baselines}

To demonstrate the effectiveness of \model{}, we compare it against a variety of state-of-the-art baselines. These baselines can be broadly categorized into three groups based on their underlying retrieval and reasoning architectures:

\paragraph{\textbf{Base MLLMs.}}
We utilize a Vision-Language Model as our primary backbone for both graph construction (e.g., OpenIE and visual parsing) and the final downstream generation process.
\begin{itemize}[leftmargin=*]
    \item \textbf{Qwen3-VL-8B}: An open-source Multimodal Large Language Model demonstrating strong visual perception and logical reasoning capabilities. We evaluate it under a zero-shot setting to establish the intrinsic reasoning lower bound of the generative backbone without external retrieval augmentation. \url{https://huggingface.co/Qwen/Qwen3-VL-8B-Instruct}
\end{itemize}

\paragraph{\textbf{Multimodal Vector-Matching RAG.}}
These baselines represent the latent space matching paradigms. They focus on mapping multimodal queries and heterogeneous documents into a shared continuous embedding space to perform dense semantic retrieval.
\begin{itemize}[leftmargin=*]
    \item \textbf{VLM2Vec-V2.0}: A unified multimodal embedding framework built upon a VLM backbone. \url{https://huggingface.co/VLM2Vec/VLM2Vec-V2.0}
    \item \textbf{VisRAG}: A vision-centric retrieval-augmented generation paradigm that processes document pages directly as images.  \url{https://github.com/openbmb/visrag}
    \item \textbf{CoRe-MMRAG}: A collaborative retrieval framework tailored for complex multi-hop reasoning over multimodal evidence. \url{https://github.com/iLearn-Lab/ACL25-COREMMRAG}
    \item \textbf{ViDoRAG}: A coarse-to-fine visual document retrieval framework that employs a Gaussian Mixture Model (GMM) for multimodal hybrid retrieval. It utilizes a sophisticated multi-agent iterative workflow to deeply process visually rich documents. \url{https://github.com/Alibaba-NLP/ViDoRAG}
\end{itemize}

\paragraph{\textbf{Graph-Enhanced RAG.}}
These models, acting as the most direct competitors to our method, introduce structural and topological modeling to explicitly enhance multi-hop reasoning and overcome the limitations of isolated vector matching.
\begin{itemize}[leftmargin=*]
    \item \textbf{HM-RAG}: A hierarchical multi-agent multimodal RAG framework designed for complex queries. It explicitly decomposes queries and performs parallel, modality-specific retrieval across heterogeneous data ecosystems, subsequently fusing the evidence via a dedicated decision agent. \url{https://github.com/ocean-luna/HMRAG}
    \item \textbf{MMGraphRAG}: A multimodal GraphRAG system that bridges vision and language by constructing a unified Multimodal Knowledge Graph (MMKG). By explicitly extracting textual entities and visual scene graphs, and aligning them via cross-modal entity linking, it enables structural reasoning paths across modalities. \url{https://github.com/wanxueyao/mmgraphrag}
\end{itemize}

\section{Implementation Details}
\label{sec:appendix_impl}

In this section, we provide comprehensive implementation details of \model{} and the specific configurations used for the baselines to ensure reproducibility.

\subsection{Implementation Details of DualG-MRAG}
\label{sec:impl_model}

All experiments for our proposed \model{} are conducted on a single NVIDIA A100 GPU. The key hyperparameters and foundation models utilized in our framework are summarized in Table~\ref{tab:hyperparams}. 

Furthermore, for ScienceQA, the candidate contexts typically consist of a pre-aligned text snippet and an optional image. Since this eliminates the need for cross-modal fusion, we simply calculate the final ranking score for this dataset by directly adding the macro-topological and micro-matching scores. 

\begin{table}[h]
\centering
\caption{Key implementation details and hyperparameters of \model{}.}
\label{tab:hyperparams}
\setlength{\tabcolsep}{6.2mm}
\begin{tabular}{lc}
\toprule
\textbf{Parameter / Configuration} & \textbf{Value} \\
\midrule
\multicolumn{2}{l}{\textit{Foundation Models \& Encoders}} \\
OpenIE \& Vision Parser & Qwen3-VL-8B \\
Text Encoder & all-mpnet-v2 \\
Semantic Linking Encoder & ColBERT \\
\midrule
\multicolumn{2}{l}{\textit{Graph Construction \& Retrieval}} \\
Entity Resolution Threshold ($\tau$) & 0.8 \\
Top-$K$ for Subgraph Search & 3 \\
Text Embedding Dimension & 768 \\
\midrule
\multicolumn{2}{l}{\textit{GNN Retriever \& Path Decoding}} \\
Message Passing Layers ($L$) & 6 \\
GNN Hidden Dimension & 512 \\
Visual Score Weight ($\alpha$) & 0.1 \\
Micro-matching Weight ($\beta$) & 0.9 \\
Unverified Decay Scale & 0.25 \\
Flow Temperature ($\tau_{\text{flow}}$) & 1.0 \\
\bottomrule
\end{tabular}
\end{table}

\subsection{Experimental Setup and Baseline Configurations}
\label{sec:impl_baselines}

To guarantee a fair and rigorous comparison, we enforce a unified system prompt across all methods during the final answer generation phase. Additionally, we apply a few specific settings tailored to the datasets and baselines. For WebQA, since a portion of the queries requires fewer reasoning hops, we simply prompt the MLLM to assess query complexity and selectively bypass path injection for simpler questions to maintain generation flexibility. For ScienceQA, to mitigate potential noise from retrieved contexts across all baselines, we instruct the MLLM to generate two candidate answers (with and without retrieval) and select the more logical one based on its internal knowledge. Finally, to maintain a strictly controlled environment where all models rely solely on the provided local corpus, we disable the external web search functionality originally included in the HM-RAG baseline.

\clearpage 

\begin{figure*}[htpb]

    \section{Case Study}
    \label{sec:appendix_case}
    \vspace{1em}

    \centering
    \includegraphics[width=1\linewidth]{figure/01.pdf} 
    \caption{\textbf{Visualization of the DualG-MRAG Pipeline.} Our framework isolates fine-grained visual verification (\textit{Micro Matching}),such as grounding the "yellow painted section" to specific candidate images,from abstract logical routing (\textit{Macro Reasoning}). The verified local anchors activate the macro-graph, where a query-driven GNN propagates relevance signals across cross-document entities. Ultimately, the system decodes the message-passing trajectory into an explicit structured reasoning path. Guided by this topological chain, the downstream generative model successfully deduces the correct answer.}
    \label{fig:case_pipeline}
    
    \vspace{1em}

    \centering
    \includegraphics[width=1\linewidth]{figure/02.pdf}
    \caption{\textbf{Comparison of Retrieved Contexts and Generated Answers.} This case dissects a cross-modal reasoning query to highlight the indispensability of explicit structural guidance.}
    \label{fig:case_path}
\end{figure*}

\end{document}